%% file: main.tex
\documentclass[10pt,twocolumn,letterpaper]{article}

\usepackage{multirow}
\usepackage{pifont}
\usepackage{wacv}              

\usepackage[accsupp]{axessibility}

\definecolor{wacvblue}{rgb}{0.21,0.49,0.74}
\usepackage[pagebackref,breaklinks,colorlinks,allcolors=wacvblue]{hyperref}

\def\wacvPaperID{128} 
\def\confName{WACV}
\def\confYear{2027}

\title{CRAFT-E: An Interpretable and Editable Neuro-Symbolic Framework for Affordance-Grounded Active Object Retrieval}

\author{Zhou Chen, Joe Lin, Carson Bulgin, Sathyanarayanan N. Aakur\\
CSSE Department, Auburn University\\
Auburn, AL, USA 36849\\
{\tt\small \{zzc0053,jzl0277,ccb0082,san0028\}@auburn.edu}
}

\begin{document}
\maketitle
\input{sec/0_abstract}    
\input{sec/1_intro}

\input{sec/2_realated_work}
\input{sec/3_method}
\input{sec/4_experiment}
\input{sec/5_evaluation}
\input{sec/6_conclusion}
{
    \small
    \bibliographystyle{ieeenat_fullname}
    \bibliography{main}
}

\end{document}

%% file: sec/0_abstract.tex
\begin{abstract}
Assistive robots operating in unstructured environments must reason not only about what objects are, but how they can be used, including non-canonical and improvised uses. This requires grounding language-based action queries to objects that afford the requested function, are visually present, and can be physically retrieved. Existing approaches often rely on black-box models, fixed affordance labels, or direct object predictions, limiting transparency and controllability. 
We introduce CRAFT-E\footnote{Project page with videos and code can be found \href{https://zzc0053.github.io/CRAFT-E-page/}{here}}, a modular neuro-symbolic framework that composes an LLM-derived verb--property--object knowledge graph with CLIP-based visual alignment and energy-based grasp reasoning. The system exposes interpretable grounding paths, incorporates grasp feasibility into affordance inference, and supports targeted graph edits without retraining. We also construct a benchmark dataset with unified annotations for verb--object compatibility, segmentation, and grasp candidates, and deploy the full pipeline on a physical robot. 
Across static scenes, real-world trials involving 20 verbs and 39 objects, ImageNet-based functional retrieval, and COCO-Tasks transfer, CRAFT-E provides a competitive training-free prior for affordance-grounded retrieval while exposing component-level diagnostics and graph-level interventions. Under staged and full-pipeline perception, CRAFT-E matches strong flat LLM priors in several settings while adding failure attribution and editable reasoning paths. 
By coupling symbolic reasoning with embodied perception, CRAFT-E offers an inspectable and customizable framework for affordance-grounded object retrieval. 

\end{abstract}

%% file: sec/1_intro.tex
\section{Introduction}
\label{sec:intro}

\begin{figure}[t]
    \centering
    \includegraphics[width=\linewidth]{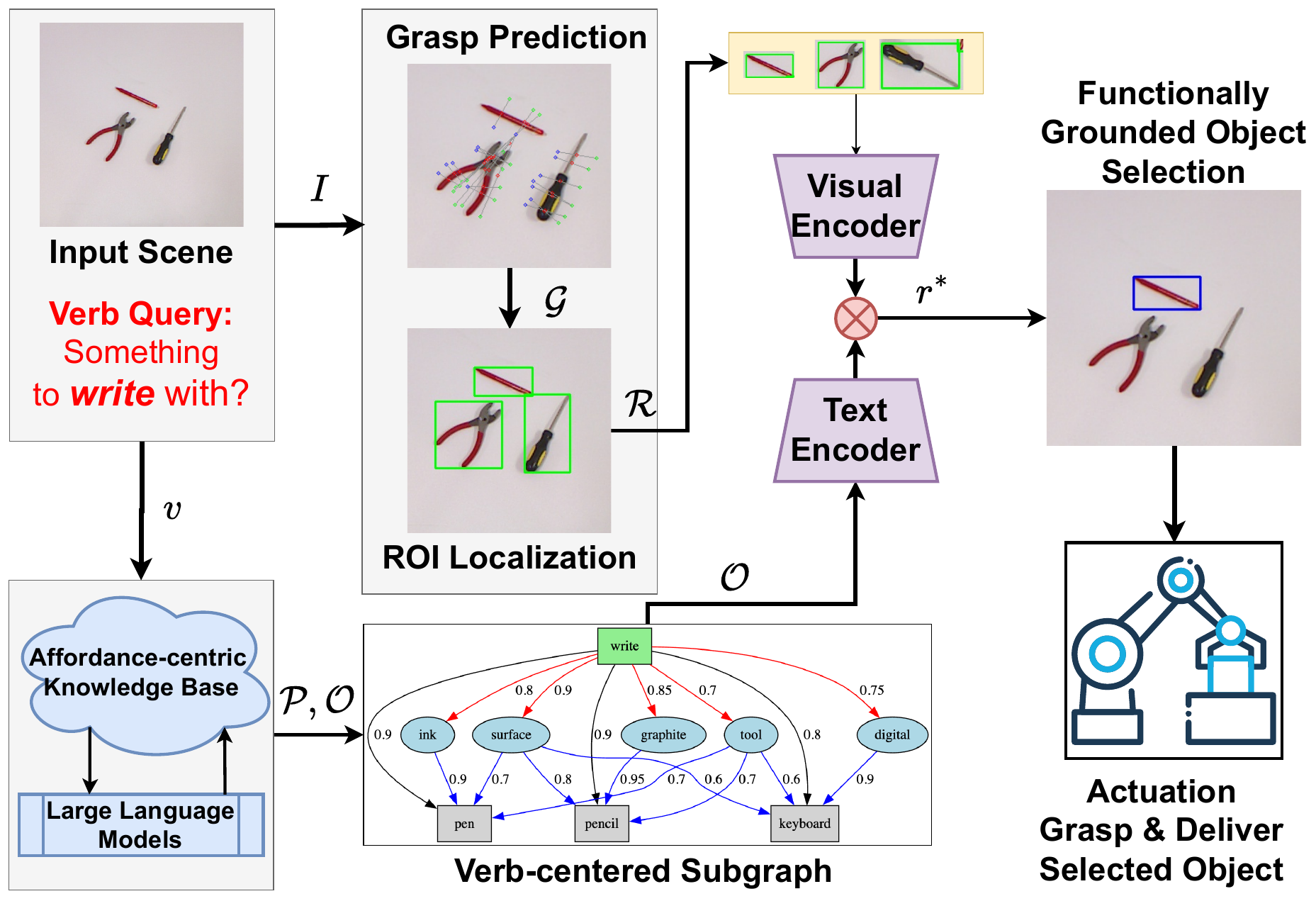}
    \caption{\textbf{Overview.} Given a scene and verb query, CRAFT-E predicts graspable candidate regions, reasons over an affordance-centric verb subgraph, and uses frozen vision-language alignment to select a functionally appropriate and retrievable object.}
    \label{fig:framework}
\end{figure}

Assistive robots operating in homes, clinics, and workplaces must respond to human requests that are functional rather than categorical. A user may ask for ``something to write with,'' ``something to cut with,'' or ``something to scoop with,'' rather than an object by its dataset label. Fulfilling such requests requires the robot to infer the intended function, identify which visible object affords that function, and determine whether it can be physically retrieved. We refer to this problem as \emph{affordance-grounded active object retrieval}: given a visual scene and a verb-level functional query, the system must select and retrieve an object that supports the requested action. This task is especially important in assistive settings, where decisions must be accurate, transparent, inspectable, and customizable~\cite{cantucci2020towards,lewis2018role,campagna2025systematic}.

Existing affordance, detection, and vision-language tasks capture important parts of this problem, but stop short of affordance-grounded retrieval. Traditional affordance and manipulation methods typically assume that the object of interest is known, and estimate how to act on it through grasp poses, 6D pose, contact regions, keypoints, or functional parts~\cite{myers2015affordance,varley2017shape,manuelli2019kpam,katz2014perceiving,fitzgerald2021modeling,abelha2017learning}. Functional affordance and task-oriented detection methods ask which objects or categories are relevant to an action; benchmarks such as COCO-Tasks~\cite{sawatzky2019object} evaluate task-relevant localization, but stop before physical interaction. Open-vocabulary detectors~\cite{joseph2021towards} broaden recognizable categories, but do not determine whether a specific object is functionally appropriate, graspable, and retrievable. Recent LLMs, VLMs, and VLAs can produce plausible choices or action plans from language prompts~\cite{team2023gemini,hurst2024gpt,radford2021learning,chen2024taskclip,zitkovich2023rt,kim2024openvla}, but their functional reasoning is often not exposed as an editable structure. Existing functional grounding benchmarks largely evaluate static verb--object compatibility, while VLA evaluations test full policies whose internal affordance reasoning is difficult to isolate. This gap motivates a new benchmark and evaluation protocol where success requires connecting a verb-level request to a specific visible object, exposing the reasoning behind that choice, and verifying that the object can be physically acted upon.

To address this challenge, we introduce \textit{CRAFT-E}, an interpretable and editable neuro-symbolic framework for affordance-grounded active object retrieval. CRAFT-E is built around a simple principle: an embodied affordance decision should expose the functional assumptions that connect a user request to a physical object. As shown in Figure~\ref{fig:framework}, CRAFT-E represents each query through an affordance-centric verb--property--object graph, where intermediate functional properties provide explicit evidence for why an object may support the requested action. This graph serves as the reasoning interface between language, visual perception, and embodiment, allowing different affordance priors to be compared under the same grounding and grasp/actionability scaffold. The resulting energy formulation selects the object that best satisfies the requested function under both visual and physical constraints. Through its externalized reasoning structure, CRAFT-E enables component-level diagnosis and targeted edits to affordance knowledge without retraining, making it inspectable and correctable rather than a black-box object selector.

Our \textbf{contributions} are four-fold: (i) we formulate \emph{affordance-grounded active object retrieval}, unifying verb-level functional grounding with physical object retrieval; (ii) we provide a new RGB-D tabletop benchmark and evaluation protocol with verb--object compatibility, object localization, and grasp annotations for staged, end-to-end, and controlled affordance-prior evaluation; (iii) we present CRAFT-E, which externalizes affordance reasoning into an editable verb--property--object graph that connects functional language, visual evidence, and physical actionability; and (iv) we demonstrate that this explicit reasoning layer supports component-level diagnosis, targeted knowledge interventions, and competitive training-free transfer, with evaluations on our new benchmark, a real-world robotic testbed, an ImageNet-derived functional grounding benchmark, and COCO-Tasks for non-canonical object use.

%% file: sec/2_realated_work.tex
\section{Related Work}
\label{sec:related}

\textbf{Affordance grounding and task-oriented detection.}
Affordance grounding studies how visual systems infer what objects can \emph{do}, not only what they \emph{are}. Task-driven object detection first showed that object relevance depends on functional context rather than category labels alone~\cite{sawatzky2019object}. Recent methods use LLMs and VLMs to extract task- or affordance-centric knowledge, grounding detection in semantic attributes, task descriptions, or implicit functional priors~\cite{tang2023cotdet,chen2024taskclip,cuttano2024clip,jang2024intra,tang2025uad}. Benchmarks such as COCO-Tasks~\cite{sawatzky2019object} formalize task-relevant localization by asking models to identify objects associated with a task. However, they often stop at prediction or localization: they do not require selecting a specific visible object for retrieval, verifying grasp feasibility, or editable reasoning traces.

\textbf{Embodied affordances and manipulation.}
Geometric, part-based, and manipulation-centered affordance methods estimate how an object can be acted upon using contact regions, grasp poses, 6D pose, keypoints, or functional parts~\cite{myers2015affordance,varley2017shape,manuelli2019kpam,katz2014perceiving,fitzgerald2021modeling,abelha2017learning}. Language-conditioned and 3D affordance grounding extend this direction toward instruction-following and embodied interaction~\cite{ju2024robo,zhu2025grounding}, while recent VLMs and VLAs can generate plausible object choices or action plans from prompts~\cite{team2023gemini,hurst2024gpt,radford2021learning,zitkovich2023rt,kim2024openvla}. These approaches are powerful, but typically assume a specified target object or collapse perception, functional reasoning, and action selection into opaque end-to-end predictions. In contrast, affordance-grounded active object retrieval isolates affordance priors under fixed grounding, grasping, and reachability constraints, requiring the system to identify which visible object satisfies a verb-level request, whether it is retrievable, and why that choice was made.

\textbf{Neuro-symbolic and graph-based reasoning.}
Neuro-symbolic approaches combine neural perception with structured representations for compositional and interpretable inference. Prior work has used LLMs, VLMs, and knowledge graphs for commonsense reasoning over objects, attributes, and actions~\cite{tang2023cotdet,chen2024taskclip,10.1007/978-3-319-10605-2_27}. CRAFT~\cite{chen2025craft} uses ConceptNet-derived commonsense paths for verb-conditioned affordance grounding, but it evaluates functional compatibility rather than physically actionable retrieval, and general-purpose relations can introduce ambiguity and semantic drift. Broader graph representations, including activity graphs, action scene graphs, dynamic scene graphs, scene graphs, visual relationship graphs, and functional 3D scene graphs, model objects and relations for recognition, planning, or relational grounding~\cite{toumpa2023object,7041337,rodin2024action,wang2024oed,kundu2023ggt,jiang2025enhancing,salzmann2024scene,zhang2025open}. FOON represents object-action dependencies through functional units for procedural manipulation reasoning~\cite{8460200,7759413}, but focuses on curated task knowledge rather than visually grounded, verb-conditioned retrieval. 
Instead, CRAFT-E embeds a verb--property--object graph in a shared retrieval scaffold for controlled ablation, diagnosis, and graph edits. 

%% file: sec/3_method.tex
\section{Proposed Approach: CRAFT-E}

\textbf{Overview.}
Given an RGB-D observation $I$ of a cluttered scene and a verb-level query $v$, the goal is to select an object that affords the requested function and can be physically retrieved. Let $\mathcal{R} = \{ r_1, \ldots, r_N \}$ denote the set of candidate regions of interest (ROIs) in $I$. Each ROI $r_i$ is defined by a 2D bounding box $\mathbf{b}_i = (x_i, y_i, w_i, h_i)$, an object hypothesis $o_{r_i}$, and a set of associated 3D grasp poses $\mathcal{G}_i = \{ \mathbf{g}_{i1}, \ldots, \mathbf{g}_{iM_i} \}$, where $\mathbf{g}_{ij} \in SE(3)$ encodes an end-effector pose. CRAFT-E scores each candidate by combining functional affordance reasoning, visual-language grounding, and physical actionability. For each ROI $r_i$, we define the total energy as
\begin{equation}
E(v, r_i) =
E_{\text{grasp}}(\mathcal{G}_i)
+ E_{\text{aff}}(v, o_{r_i})
+ E_{\text{align}}(\mathbf{b}_i, v),
\label{eqn:overall_energy}
\end{equation}
where $E_{\text{grasp}}$ denotes physical actionability, penalizing grasps that are unreliable, unreachable, or infeasible from the robot's current configuration; $E_{\text{aff}}$ measures how strongly the object hypothesis $o_{r_i}$ supports the requested action $v$; and $E_{\text{align}}$ measures visual--linguistic alignment between the ROI and the query. 
All terms are normalized over the candidate ROIs in a scene to a comparable range. Object retrieval is then selecting the lowest-energy ROI:
\begin{equation}
r^* = \arg\min_{r_i \in \mathcal{R}} E(v, r_i),
\label{eqn:overall_optimization}
\end{equation}
thus making each decision decomposable into physical actionability, affordance reasoning, and visual grounding.

\begin{figure}[t]
    \centering
    \includegraphics[width=0.99\linewidth]{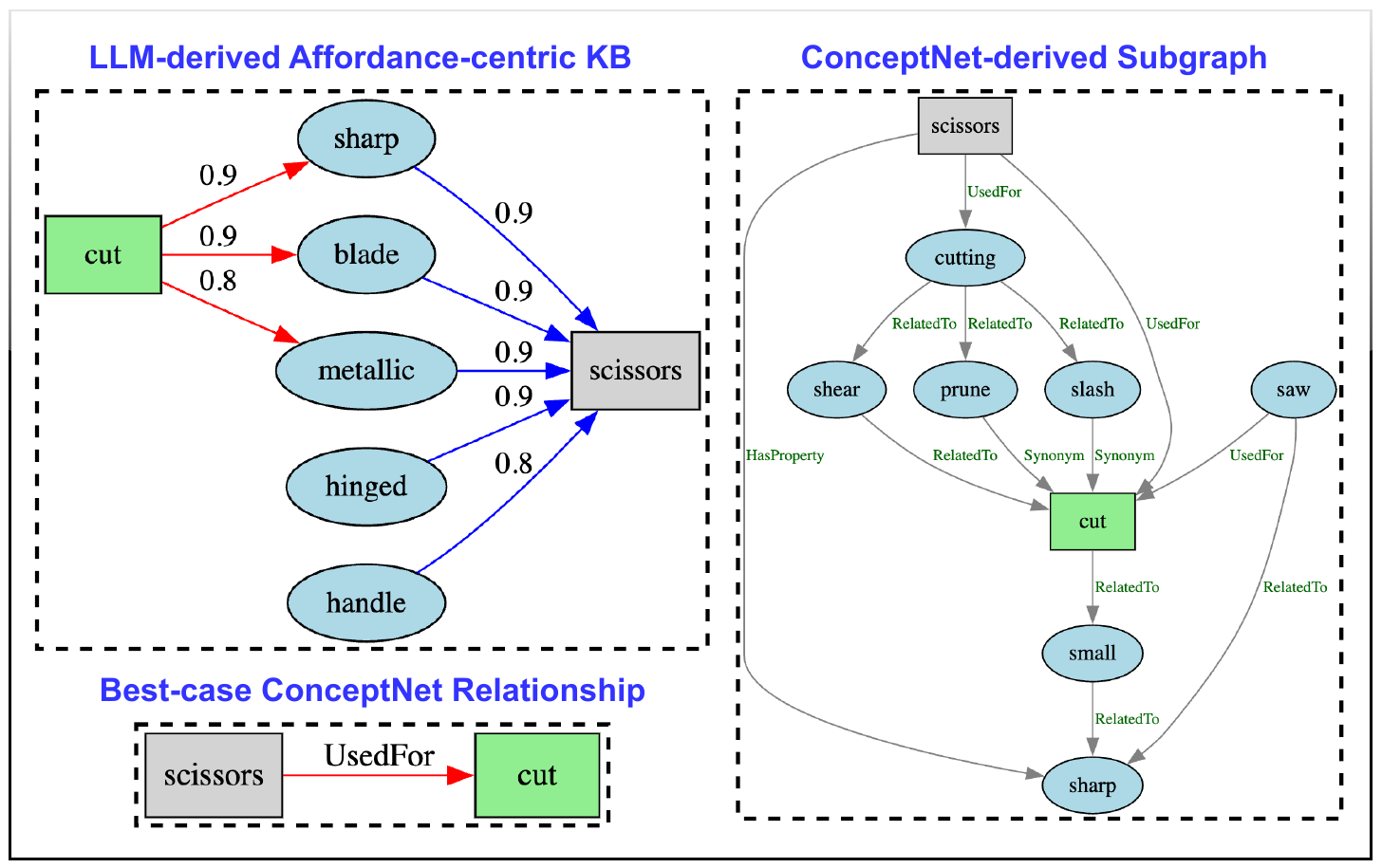}
    \caption{LLM-derived affordance knowledge (left) versus ConceptNet subgraphs (right). CRAFT-E links verbs to objects through interpretable functional properties, while ConceptNet relies on general-purpose relations such as \texttt{RelatedTo}, which can yield noisy paths with limited explanatory structure.}
\label{fig:llm_vs_conceptnet}
\end{figure}

\subsection{Affordance Hypothesis Generation}

Given a query verb $v$ (e.g., \textit{cut}, \textit{scoop}, \textit{write}), we begin by estimating which object hypotheses are functionally compatible with the requested action. Rather than relying on static knowledge graphs such as ConceptNet~\cite{speer2017conceptnet} or fixed verb-category mappings~\cite{nguyen2020robot,nguyen2022affordance}, we construct a task-specific affordance-centric knowledge base (KB) using large language models (LLMs)~\cite{team2023gemini,hurst2024gpt}. The KB is generated offline as a bipartite verb--property--object graph. For each verb $v$, an LLM produces a set of descriptive properties $\mathcal{P}_v = \{ p_1, \ldots, p_k \}$, such as \texttt{sharp}, \texttt{blade}, or \texttt{metallic} for \texttt{cut}, with relevance weights $w_{vp} \in [0,1]$. For each object $o \in \mathcal{O}$, the LLM then scores how well $o$ satisfies each property $p \in \mathcal{P}_v$, yielding weights $w_{po} \in [0,1]$. This produces interpretable two-hop paths $v \rightarrow p \rightarrow o$ without invoking LLMs at inference time, enabling low-latency deployment.

Figure~\ref{fig:llm_vs_conceptnet} compares this affordance-centric KB with ConceptNet-derived alternatives. ConceptNet can recover surface-level relations such as \textit{scissors} \texttt{UsedFor} \textit{cut}, but its general-purpose relations (e.g., \texttt{RelatedTo}, \texttt{Synonym}, \texttt{HasProperty}) often introduce semantic drift and noisy non-functional paths. In contrast, our KB emphasizes task-relevant properties that are actionable and often visually verifiable through alignment models. Unlike direct LLM prompting at inference time, even with reasoning traces~\cite{xu2025towards} or scratchpads~\cite{hsu2023can}, the explicit graph exposes how an object satisfies a verb through intermediate properties. This makes the reasoning inspectable, editable, and compatible with visual alignment models such as CLIP~\cite{radford2021learning}, while allowing nodes or edges to be swapped or modified without retraining.

We score the plausibility of object $o$ affording verb $v$ by aggregating weighted paths in the KB:
\begin{equation}
E_{\text{aff}}(v, o) =
\tanh \left(
- \sum_{p \in \mathcal{P}_v(o)} \left( w_{vp} + w_{po} \right)
\right),
\label{eqn:affordance_energy}
\end{equation}
where $\mathcal{P}_v(o)$ is the query-conditioned subset of properties connecting $v$ to $o$; unrelated object neighbors or graph degree do not affect $E_{\text{aff}}$. Summation allows multiple matched task-relevant properties to provide stronger evidence of functional compatibility than a single path. Lower energy indicates stronger compatibility, and the $\tanh$ activation provides bounded, smooth scores without global normalization. The same formulation can integrate different KB sources, including LLM-derived graphs, ConceptNet-based graphs as in CRAFT~\cite{chen2025craft}, or hand-crafted knowledge, and supports reasoning over unlabeled candidate regions when the relevant functional concepts are represented in the KB. 

\subsection{Visual Grounding via ROI-Level Alignment}
The affordance graph provides a functional compatibility score for each object hypothesis, but symbolic plausibility alone does not ensure that the corresponding candidate is visually present or correctly localized in the scene. We therefore ground each candidate ROI $r_i \in \mathcal{R}$ against the query verb $v$ using a pretrained vision-language model. Specifically, we extract an image embedding $\phi_I(r_i)$ from the CLIP~\cite{radford2021learning} vision encoder for the crop defined by $\mathbf{b}_i$, and encode the verb query using the CLIP text encoder as $\phi_T(v)$. The visual alignment energy is then defined as
\begin{equation}
E_{\text{align}}(\mathbf{b}_i, v) =
\tanh\left(
-\cos\left(\phi_I(r_i), \phi_T(v)\right)
\right),
\label{eqn:visual_alignment_energy}
\end{equation}
where $\cos(\cdot,\cdot)$ denotes cosine similarity. Lower energy indicates stronger visual--linguistic agreement, and the $\tanh$ transformation yields bounded, smooth scores. 
CRAFT-E uses ROI-level alignment: each full candidate region is compared directly with the action query, rather than separately scoring intermediate properties such as \textit{sharp} or \textit{hinged}. This keeps visual grounding lightweight and allows the same alignment module to be reused across knowledge sources and evaluation settings. 
Importantly, the goal is not to assign semantic labels to ROIs, but to determine which visually grounded region best supports the functional hypothesis produced by the affordance graph. This enables functional grounding over unlabeled or ambiguous candidate ROIs without requiring class-level object recognition.

\subsection{Physical Actionability via Grasp Reasoning}

The final term ensures that the selected object is not only functionally compatible and visually grounded, but also physically retrievable by the robot. Each candidate ROI $r_i$ is associated with a set of predicted 6-DOF grasp poses $\mathcal{G}_i = \{ \mathbf{g}_{i1}, \ldots, \mathbf{g}_{iM_i} \}$. We use the confidence scores returned by a pretrained grasp prediction network to estimate physical actionability, which captures whether an object can be reliably grasped and retrieved from the robot's current configuration. Formally, we define
\begin{equation}
E_{\text{grasp}}(\mathcal{G}_i) =
- \log \left(
\max_{\mathbf{g}_{ij} \in \mathcal{G}_i} s_{ij}
\right),
\label{eqn:grasp_energy}
\end{equation}
where $s_{ij} \in (0,1]$ is the confidence score for grasp pose $\mathbf{g}_{ij}$. Lower energy indicates higher physical actionability. Before summation, $E_{\text{grasp}}$ is normalized over candidate ROIs to match the scale of $E_{\text{aff}}$ and $E_{\text{align}}$. This term prioritizes objects with at least one reliable grasp and penalizes candidates that are difficult to retrieve due to poor grasp quality, reachability, or execution constraints. 
Together, $E_{\text{aff}}$, $E_{\text{align}}$, and $E_{\text{grasp}}$ connect the three requirements of affordance-grounded active object retrieval: functional compatibility, visual grounding, and physical actionability. Because each term is computed separately, the final decision can be inspected at the component level, enabling diagnosis of whether failures arise from affordance reasoning, visual grounding, or embodied execution.

\begin{table}[t]
\centering
\small
\resizebox{\columnwidth}{!}{
\begin{tabular}{llccc}
\toprule
\textbf{Grasp Model} & \textbf{Affordance Prior} & \textbf{3D (\%)} & \textbf{Detect (\%)} & \textbf{Grasp Rate (\%)} \\
\midrule
\multirow{4}{*}{GraspNet}
& Gemini  & 83.89 & \textbf{56.67} & 44.44 [32.2, 57.2] \\
& GPT-4o  & 82.22 & 52.22 & 46.11 [33.3, 58.9] \\
& CRAFT   & \textbf{86.11} & 37.78 & 33.33 [25.0, 43.3] \\
& CRAFT-E & 85.00 & 52.22 & \textbf{46.67} [33.9, 60.0] \\
\midrule
\multirow{4}{*}{KpNet}
& Gemini  & 72.78 & 45.00 & 23.89 [17.2, 30.6] \\
& GPT-4o  & \textbf{78.89} & 45.00 & 33.33 [27.8, 38.9] \\
& CRAFT   & 78.33 & 33.89 & 25.56 [19.4, 31.7] \\
& CRAFT-E & 77.22 & \textbf{51.67} & \textbf{40.00} [30.0, 50.0] \\
\bottomrule
\end{tabular}
}
\caption{\textbf{Real-world evaluation} of 3D grasp feasibility, detection accuracy, and grasp execution success. 
Only \(E_{\mathrm{aff}}\) is varied; the retrieval scaffold is fixed. 
Grasp Rate includes 95\% paired-bootstrap confidence intervals over 20 verb-object pairs.}
\label{tab:real_world_results}
\end{table}

\subsection{Implementation Details} 
CRAFT-E integrates off-the-shelf pretrained models without task-specific training. We use GraspNet~\cite{mousavian20196}, pretrained on GraspNet-1Billion~\cite{fang2020graspnet}, and GraspKpNet~\cite{xu2022gknet}, pretrained on the Cornell dataset, to generate 6-DOF grasp candidates and obtain uncalibrated grasp confidence scores. In physical trials, grasp candidates are further constrained by reachability and execution feasibility from the robot's current configuration. Object regions are segmented using Segment Anything Model v2 (SAMv2)~\cite{ravi2024sam}, with predicted grasp locations used as prompts, and bounding boxes extracted as candidate ROIs. 
Visual embeddings are computed using CLIP~\cite{radford2021learning} with a ViT-B/32 backbone~\cite{dosovitskiyimage}, applied to $224 \times 224$ ROI crops; verb queries are encoded with the corresponding CLIP text encoder. Affordance reasoning uses a symbolic KB constructed offline with Gemini 2.5 Flash~\cite{team2023gemini}, which generates verb--property and property--object associations with 10 properties per verb. At inference, the system selects the lowest-energy (from Eq.~\ref{eqn:overall_energy}) candidate ROI. All components are run on a workstation with an RTX 3060TI GPU, averaging 0.62s per scene.

%% file: sec/4_experiment.tex
\section{Experimental Setup}


\begin{table}[t]
\centering
\resizebox{\columnwidth}{!}{%
\begin{tabular}{c|c|c|l|c}
\toprule
\textbf{Setting} & \textbf{Grasp} & \textbf{ROI} & \textbf{Affordance Prior} & \textbf{Accuracy (\%)} \\
\midrule
\multirow{5}{*}{\shortstack{\textbf{Perfect}\\\textbf{Grasp \& Detect}}}
  & \multirow{4}{*}{GT}         & \multirow{4}{*}{GT}    & Gemini            & 59.79 \\
  &         &    & GPT-4o            & 62.89 \\
  &         &    & {CRAFT}       & {58.76} \\
  &         &    & {CRAFT-E}       & \textbf{64.95} \\
  &         &    & CRAFT-E+GPT-4o   & 52.58 \\
\midrule
\multirow{4}{*}{\textbf{Perfect Grasp}} 
  & GT        & \multirow{4}{*}{SAMv2}      & Gemini            & 56.70 \\
  & GT        &      & {GPT-4o}   & \textbf{59.79} \\
  & GT        &      & CRAFT          & 52.58 \\
  & GT        &      & CRAFT-E          & 56.70 \\
  & GT        &      & CRAFT-E+GPT-4o   & 46.39 \\
\midrule
\multirow{8}{*}{\textbf{Full Pipeline}} 
  & GraspNet  & \multirow{4}{*}{SAMv2}      & Gemini            & 19.59 \\
  & GraspNet  &      & {GPT-4o}   & \textbf{20.62} \\
  & GraspNet  &      & CRAFT          & {12.37} \\
  & GraspNet  &      & CRAFT-E          & \textbf{20.62} \\
  & GraspNet  &      & CRAFT-E+GPT-4o   & 16.49 \\
\cmidrule{2-5}
  & KpNet     & \multirow{4}{*}{SAMv2}     & Gemini            & 52.58 \\
  & KpNet     &      & {GPT-4o}   & \textbf{55.67} \\
  & KpNet     &      & CRAFT          & {49.48} \\
  & KpNet     &      & CRAFT-E          & \textbf{55.67} \\
  & KpNet     &      & CRAFT-E+GPT-4o   & 46.39 \\
\bottomrule
\end{tabular}
}
\caption{\textbf{Static evaluation} accuracy under perfect grounding, perfect grasping, and full pipeline settings. 
Only \(E_{\mathrm{aff}}\) is varied; the retrieval scaffold is fixed. 
CRAFT-E+GPT-4o uses GPT-4o to derive the LLM-KG.
}
\label{tab:static_eval_results}
\end{table}

\textbf{Static Evaluation Dataset.}
To evaluate affordance-grounded retrieval under controlled perceptual assumptions, we adapt the 97 cluttered RGB-D tabletop scenes from Chu \textit{et al.}~\cite{chu2018deep}\footnote{Original dataset is available \href{https://github.com/ivalab/grasp_multiObject}{here}}. The original dataset provides realistic multi-object scenes and grasp annotations, but lacks affordance-specific supervision. We augment it with bounding boxes and coarse labels for visible objects, a verb query specifying the functional request, and a ground-truth object that satisfies the query. Each image contains one consensus verb-object pairing, with verbs chosen to span diverse affordance types and repeated object categories assigned different verbs when possible. The final benchmark includes 97 scenes, 31 object classes, and 37 verbs; the full vocabulary is provided in the supplement.

\textbf{Real-World Evaluation.}
We also evaluate the full perception-to-action pipeline on a physical robot using 20 verb-object pairs spanning 20 verbs and 39 objects. The system is deployed on a 6-DOF Dobot CR3 arm with a 5\,kg payload and an Intel RealSense D435i RGB-D camera mounted on the arm via a custom 3D-printed bracket. Each trial contains one target object satisfying the verb query and four distractors, and the robot executes a closed-loop grasp and retrieval using the predicted ROI and grasp pose. Each verb-object pair is tested across three randomized layouts and evaluated by three human annotators, yielding 180 outcomes per affordance prior and grasp backend. Verb queries are phrased as ``give me something to \textit{\textless verb\textgreater} with.'' Annotators select objects that satisfy the intended action as tools or instruments rather than objects that only trivially support the action; disagreements are resolved by majority vote.

\textbf{Evaluation Protocol and Metrics.}
We evaluate staged and end-to-end performance. Unless otherwise stated, all affordance priors use the same ROI generation, visual alignment, grasp scoring, reachability filtering, and aggregation; only \(E_{\mathrm{aff}}\) varies. In the static setting, success requires the selected region to overlap the ground-truth box by at least 0.5 IoU and the predicted grasp to satisfy the criteria of Chu \textit{et al.}~\cite{chu2018deep}: angular deviation no greater than $30^\circ$ and Jaccard index greater than 0.25. In the real-world setting, success requires identifying the correct object region, selecting a valid reachable grasp, and transporting the object subject to inverse-kinematics constraints. These metrics jointly evaluate functional compatibility, visual localization, and physical actionability. For real-world statistics, each verb-object pair is aggregated over three layouts and three annotators and used as the paired unit. We report 95\% CIs using paired bootstrap over the 20 pairs and paired sign-flip tests for CRAFT-E vs. each baseline; layout-level pairing is not used because layouts are randomized across methods. 

%% file: sec/5_evaluation.tex
\subsection{Quantitative Results}

\begin{table}[t]
\centering
\resizebox{\columnwidth}{!}{%
\begin{tabular}{lccc}
\toprule
\textbf{Approach} & \textbf{Accuracy (\%)} & \textbf{MRR (\%)} & \textbf{nDCG (\%)} \\
\midrule
Object-Aware Oracle             & 100.00 & 69.90 & 76.80 \\
Affordance-Aware Oracle        & 73.80  & 68.70 & 82.70 \\
CLIP+verb prompting              & 38.00  & 54.80 & 56.10 \\
ResNet-RNN~\cite{nguyen2020robot,nguyen2022affordance}               & 60.20  & 65.70 & 76.50 \\
ALGO~\cite{kundu2024discovering}                     & 42.52  & 52.38 & 53.33 \\
Gemini-2.0-Flash~\cite{team2023gemini}         & 42.80  & 57.50 & 59.70 \\
GPT-4o~\cite{hurst2024gpt}                   & 45.30  & 57.80 & 60.40 \\
CRAFT~\cite{chen2025craft}                    & 44.62  & 58.20 & 61.30 \\
\textbf{CRAFT-E}  & \textbf{48.30} & \textbf{59.40} & \textbf{63.60} \\
\bottomrule
\end{tabular}
}
\caption{Functional grounding evaluation on ImageNet. Top-1 accuracy and MRR, and nDCG are reported for the single-affordance and multi-affordance settings.}
\label{tab:imagenet_eval}
\end{table}

\textbf{Static evaluation.}
Table~\ref{tab:static_eval_results} evaluates CRAFT-E under three levels of perceptual supervision: oracle grasps and detections, oracle grasps with predicted ROIs, and the full pipeline with predicted ROIs and grasps. This staged protocol separates affordance reasoning from errors introduced by segmentation and grasp prediction. Under \textit{Perfect Grasp \& Detect}, where the evaluation most directly isolates affordance reasoning, CRAFT-E achieves the highest accuracy at 64.95\%, followed by GPT-4o (62.89\%), Gemini (59.79\%), and CRAFT (58.76\%). With predicted SAMv2 regions but oracle grasps, GPT-4o performs best at 59.79\%, while CRAFT-E remains competitive at 56.70\%. In the full pipeline, performance drops as errors accumulate from ROI generation and grasp prediction; CRAFT-E matches GPT-4o with both GraspNet (20.62\%) and KpNet (55.67\%), while outperforming Gemini and CRAFT in the KpNet setting. These results show that CRAFT-E provides competitive static grounding performance while retaining an explicit structure for diagnosis and graph-level intervention. 

\textbf{Real-world evaluation.} 
Table~\ref{tab:real_world_results} reports physical robot performance. We use \textit{Grasp Rate} as the primary end-to-end metric, measuring successful verb grounding, object selection, executable grasp prediction, and retrieval. We also report \textit{3D}, whether viable grasps are produced for the correct object, and \textit{Detect}, whether the correct object patch is identified from the predicted grasp location. CRAFT-E obtains the highest Grasp Rate under both backends: 46.67\% with GraspNet and 40.00\% with KpNet. Paired sign-flip tests show significant gains over CRAFT/ConceptNet under both backends and over Gemini under KpNet; differences against GPT-4o are not significant. Thus, CRAFT-E is competitive with strong flat LLM priors while providing explicit diagnosis and graph-level intervention. Diagnostic metrics further show that the best intermediate score need not yield the best retrieval outcome: CRAFT has the highest 3D score under GraspNet (86.11\%) but the lowest Grasp Rate (33.33\%), indicating that physical actionability alone is insufficient without accurate functional grounding. 

\textbf{Functional grounding on ImageNet.}
To evaluate generalization beyond the tabletop and robot settings, we test on the functional object selection benchmark of Nguyen \textit{et al.}~\cite{nguyen2020robot,nguyen2022affordance}, which contains 216 ImageNet object categories and 50 functional verbs. Each episode pairs a verb query with candidate ImageNet images. In the single-affordance setting, exactly one image is correct and performance is measured by top-1 accuracy; in the multi-affordance setting, multiple images may be valid and performance is measured by MRR and nDCG. As shown in Table~\ref{tab:imagenet_eval}, CRAFT-E achieves the best non-oracle, training-free performance across all three metrics, with 48.30\% accuracy, 59.40\% MRR, and 63.60\% nDCG. It outperforms GPT-4o, Gemini, ALGO, and CRAFT, indicating that explicit verb--property--object reasoning improves ranking. Oracle models and ResNet-RNN (task-trained on target affordance labels) outperform CRAFT-E because they assume access to ground-truth object or affordance labels for training or inference.

\begin{table}[t]
\centering
\resizebox{\columnwidth}{!}{
\setlength{\tabcolsep}{1.0pt}
\begin{tabular}{lcccc}
\toprule
Method & Train-free & Frozen VLM & Prior & AP@50 \\
\midrule
Pick best class~\cite{sawatzky2019object} & \ding{55} & -- & freq. & 14.1 \\
Ranker~\cite{sawatzky2019object} & \ding{55} & -- & learned & 9.1 \\
Task detector~\cite{sawatzky2019object} & \ding{55} & -- & learned & 20.6 \\
Classification~\cite{sawatzky2019object} & \ding{55} & -- & learned & 28.8 \\
GGNN + fusion~\cite{sawatzky2019object} & \ding{55} & -- & learned & 32.6 \\
TOIST~\cite{li2022toist} & \ding{55} & \ding{55} & task text & 41.3 \\
TOIST$\dagger$~\cite{li2022toist} & \ding{55} & \ding{55} & task text & 44.1 \\
TaskCLIP~\cite{chen2024taskclip} & \ding{55} & partial & LLM+VLM & 48.1 \\
TaskCLIP*~\cite{chen2024taskclip} & \ding{55} & partial & LLM+VLM & 50.3 \\
CoTDet~\cite{tang2023cotdet} & \ding{55} & \ding{55} & LLM-CoT & 56.9 \\
\midrule
CLIP+verb prompting & \ding{51} & \ding{51} & none & 3.7 \\
Gemini & \ding{51} & \ding{51} & LLM objs. & 14.0 \\
GPT-5.2 & \ding{51} & \ding{51} & LLM objs. & 15.9 \\
CRAFT-E & \ding{51} & \ding{51} & LLM+KG & \textbf{21.7} \\
\bottomrule
\end{tabular}
}
\caption{\textbf{COCO-Tasks AP$_{50}$ comparison.} Train-free methods use no COCO-Tasks training labels or parameter updates. Our variants use frozen Faster R-CNN proposals and frozen CLIP features, varying only the task-to-object prior. The edited-KG result is excluded and studied separately as a knowledge intervention.}
\label{tab:cocotasks_compact}
\end{table}

\textbf{COCO-Tasks transfer.}
We further evaluate CRAFT-E on COCO-Tasks~\cite{sawatzky2019object}, a task-driven detection benchmark that emphasizes canonical and non-canonical object use. Unlike our tabletop and robot evaluations, this setting does not test embodiment or physical retrieval; it isolates visual grounding using frozen Faster R-CNN proposals and frozen CLIP features. COCO-Tasks is also distinct from pure object commonsense: for \emph{open bottle of beer}, suitable objects include chairs, dining tables, and benches as possible hard edges or leverage surfaces, while tasks such as \emph{dig hole} or \emph{pound carpet} may favor skis, surfboards, skateboards, or tennis rackets as improvised rigid tools. Table~\ref{tab:cocotasks_compact} includes supervised COCO-Tasks methods only as context, since they use task-specific labels and parameter updates. The fair comparison is among train-free variants, which share the same proposals and CLIP features and differ only in their task-to-object prior: CLIP+verb prompting uses no external prior, Gemini and GPT-5.2 use direct LLM object priors, and CRAFT-E uses the LLM-derived verb--property--object graph. CRAFT-E achieves the highest train-free AP$_{50}$ of 21.7, outperforming CLIP+verb prompting (3.7), Gemini (14.0), and GPT-5.2 (15.9), showing that structured knowledge provides a stronger prior for grounding improvised object use than direct LLM/VLM object selection.

\begin{table}[t]
\centering
\resizebox{\columnwidth}{!}{%
\begin{tabular}{l lcc}
\toprule
\textbf{Grasp Model} & \textbf{Affordance Prior} & \textbf{MRR (\%)} & \textbf{Grasp Success (\%)} \\
\midrule
\multirow{4}{*}{GraspNet}
 & Gemini     & \textbf{78.33} & \textbf{90} \\
 & GPT-4o     & 75.83 & \underline{70} \\
 & CRAFT      & 75.00 & \underline{70} \\
 & \textbf{CRAFT-E} & \underline{76.67} & \textbf{90} \\
\midrule
\multirow{4}{*}{KpNet}
 & Gemini     & \underline{80.00} & \underline{60} \\
 & GPT-4o     & \textbf{85.00} & \textbf{70} \\
 & CRAFT      & \underline{80.00} & \textbf{70} \\
 & \textbf{CRAFT-E} & \textbf{85.00} & \textbf{70} \\
\bottomrule
\end{tabular}
}
\caption{Real-world multi-affordance evaluation. MRR measures ranking quality; Grasp Success denotes successful physical execution. 
Only \(E_{\mathrm{aff}}\) is varied; the retrieval scaffold is fixed.
}
\label{tab:rw_multi_affordance}
\end{table}
\textbf{Multi-affordance ambiguity.}
Finally, Table~\ref{tab:rw_multi_affordance} evaluates real-world scenes where multiple objects may plausibly satisfy the same verb. This setting measures whether models can rank functionally valid alternatives and still produce executable retrievals. CRAFT-E remains competitive across both grasp backends, matching the best MRR under KpNet (85.00) and achieving 90\% grasp success under GraspNet. These results support the view that affordance-grounded retrieval is often a ranking problem under ambiguity rather than a single deterministic mapping from verb to object. More details and qualitative examples are in supplementary.

\begin{figure*}
    \centering
    \begin{tabular}{cccc}
    \hline
        \hline
         \multicolumn{4}{c}{\textbf{Successful Prediction (Query: ``\textit{write}'')}}\\
         \hline
         \includegraphics[width=.2\textwidth]{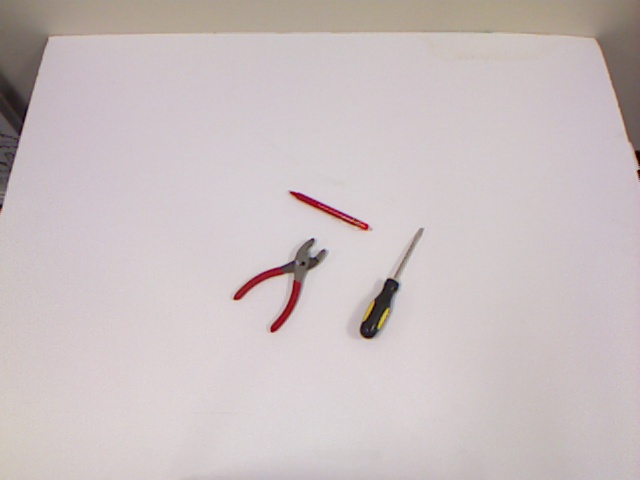} & 
         \includegraphics[width=.2\textwidth]{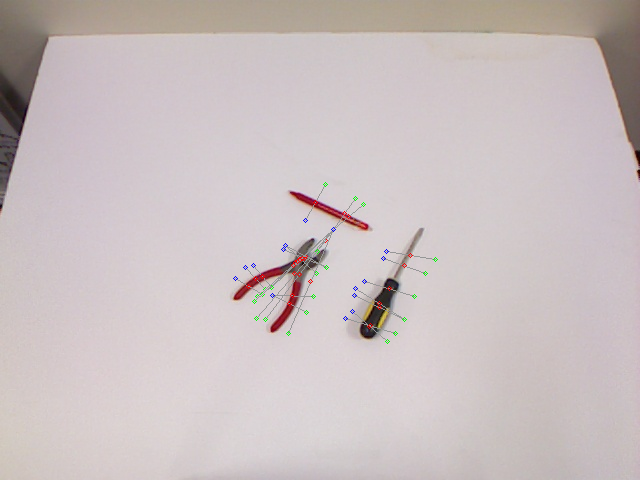} & 
         \includegraphics[width=.2\textwidth]{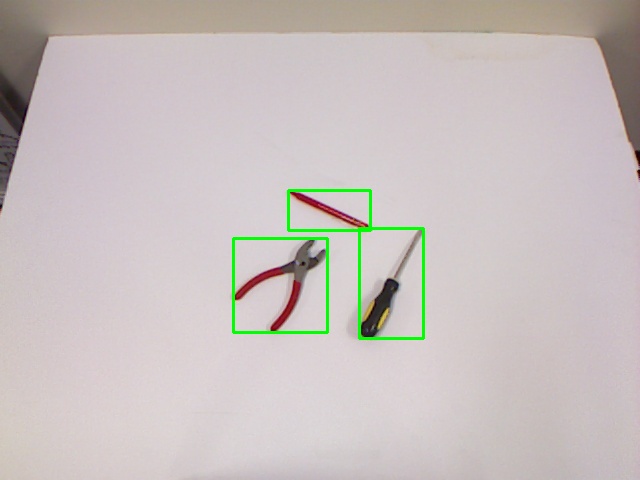} & 
         \includegraphics[width=.2\textwidth]{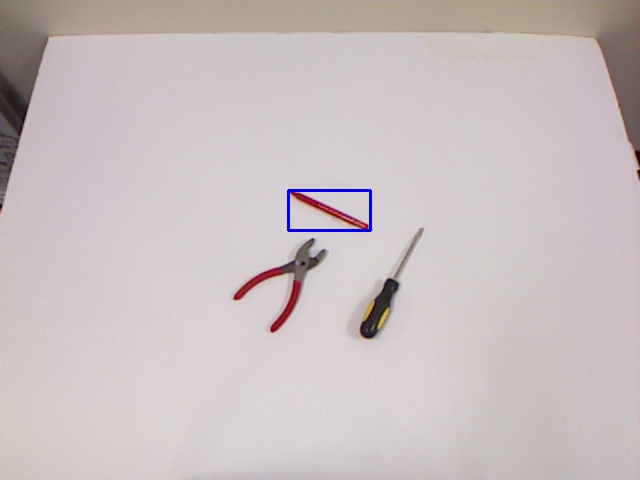} \\
         \hline
        \multicolumn{4}{c}{\textbf{Unsuccessful Prediction - Grasp Failure (Query: ``\textit{pick up}'')}}\\
         \hline
         \includegraphics[width=.2\textwidth]{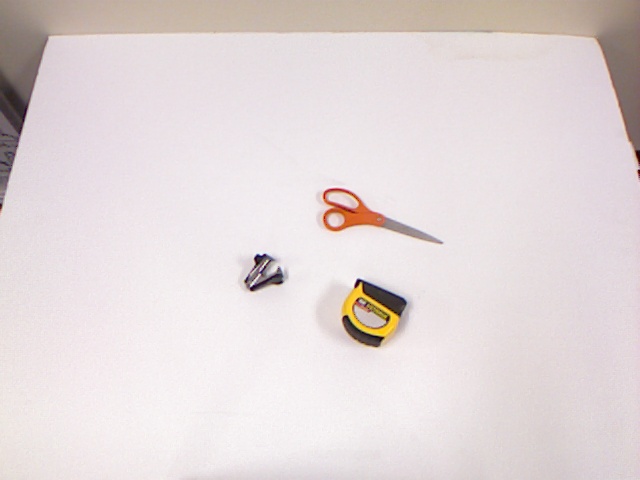} & 
         \includegraphics[width=.2\textwidth]{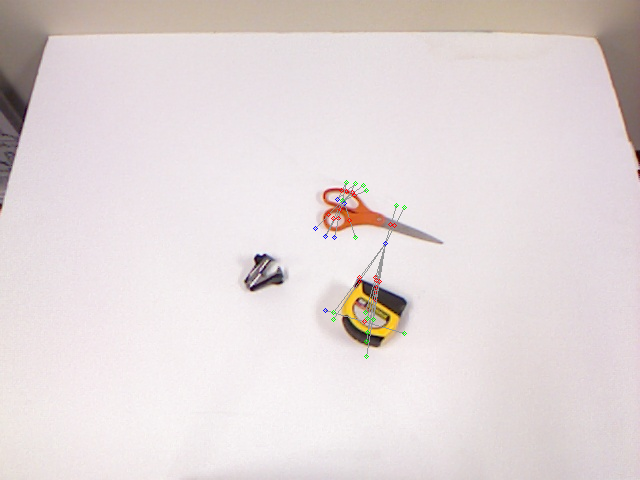} & 
         \includegraphics[width=.2\textwidth]{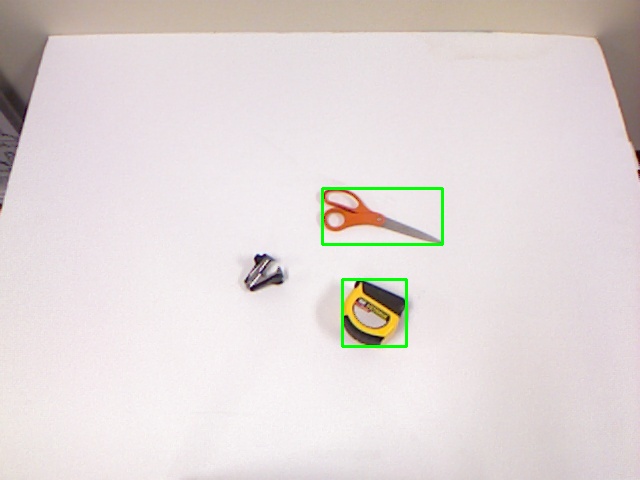} & 
         \includegraphics[width=.2\textwidth]{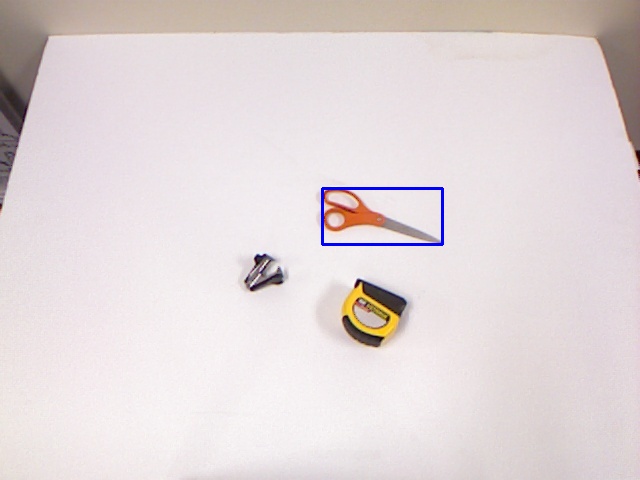} \\
         \hline
        \multicolumn{4}{c}{\textbf{Unsuccessful Prediction - Grounding Failure (Query: ``\textit{construct}'')}}\\
         \hline
         \includegraphics[width=.2\textwidth]{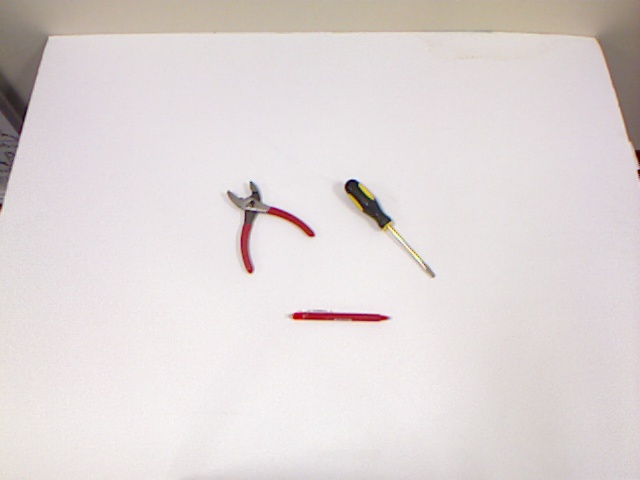} & 
         \includegraphics[width=.2\textwidth]{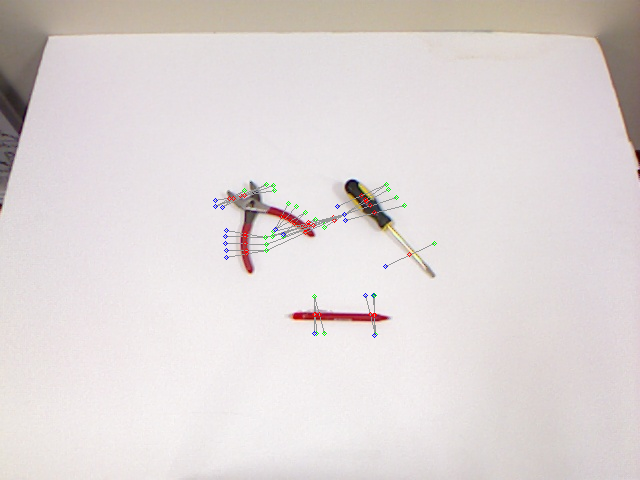} & 
         \includegraphics[width=.2\textwidth]{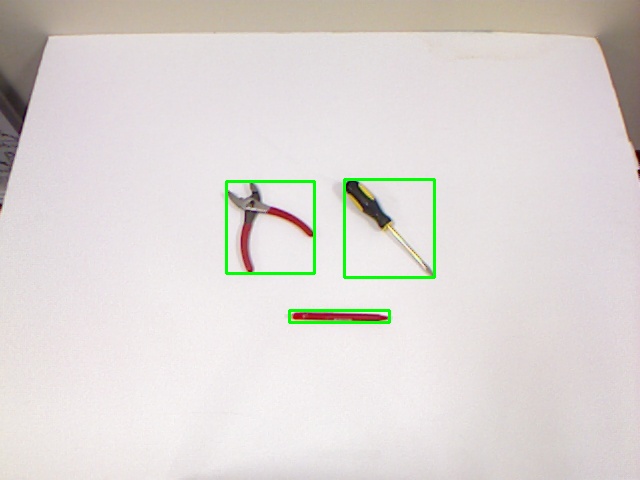} & 
         \includegraphics[width=.2\textwidth]{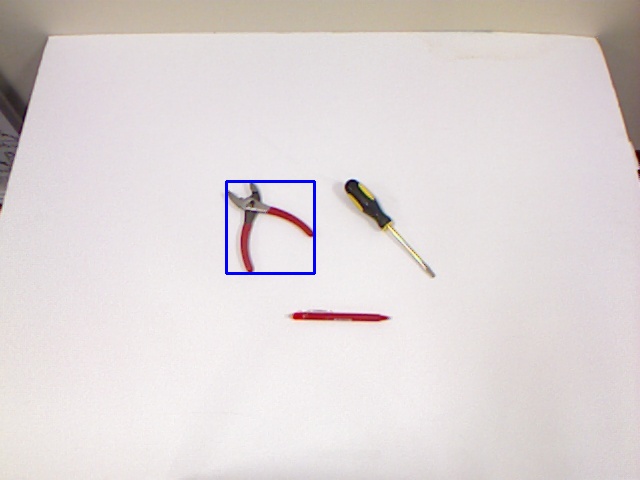} \\
    \end{tabular}
\caption{\textbf{Qualitative results} for three queries: \textit{write}, \textit{pick up}, and \textit{construct}. Columns show the input scene, grasp predictions, region proposals (green), and final grounding (blue). Top: successful retrieval. Middle: grasp failure excludes the correct object. Bottom: grounding failure selects a graspable but functionally incorrect object.}
    \label{fig:qualitative_results}
\end{figure*}

\subsection{Interpretable, Editable Affordance Grounding}

Beyond benchmark performance, we evaluate whether the affordance graph provides an interpretable and controllable reasoning layer. This intervention study tests whether failures can be traced to specific verb--property--object relations and changed without retraining perception, language, or grasping components. Starting from observed failures in the static multi-object affordance dataset, we identify noisy or weak graph elements, including overly generic properties such as \textit{body} and \textit{color}, and ambiguous property--object links such as assigning \textit{container} to objects that are not useful for containment in the queried context. We then remove or down-weight selected nodes and edges and re-evaluate the fixed pipeline. 
These edits produce same-set controllability gains: accuracy improves from 55.67\% to 60.82\% with KpNet, and Final Grasp improves from 20.62\% to 25.77\% with GraspNet. Because edits are selected after inspecting failures from the same intervention set, this is a researcher-in-the-loop controllability analysis, not held-out automatic graph repair. We also test a negative intervention for \textit{cut}: removing only \textit{sharp}$\rightarrow$\textit{scissors} does not change the top object because scissors remains supported by \textit{blade}, \textit{edge}, \textit{metallic}, and \textit{tool}. Thus, the graph is editable but not brittle to removing one redundant edge. These results show that CRAFT-E exposes decision-relevant relations that can be inspected, diagnosed, and modified directly with graph-level failure attribution and intervention without retraining. 

\subsection{Ablation Studies}

We analyze component contributions using the variants reported in Tables~\ref{tab:real_world_results} and \ref{tab:static_eval_results}. 
Compared to flat LLM priors (Gemini, GPT-4o), which predict verb--object compatibility without intermediate graph structure, CRAFT-E achieves the highest accuracy under perfect inputs (64.95\%) and remains competitive under predicted ROIs and full-pipeline perception, where differences against GPT-4o are not significant. 
CRAFT-E+GPT-4o uses GPT-4o only as the offline KG generator; its lower performance reflects sensitivity to graph construction, not a different inference pipeline. 
In real-world grasping, CRAFT-E achieves the highest Grasp Rate, with statistically supported gains over CRAFT/ConceptNet under both backends and over Gemini under KpNet, but not over GPT-4o. 
The gap between CRAFT and CRAFT-E further shows that functional reasoning must be coupled with grasp feasibility: high affordance plausibility alone does not guarantee physically executable actions. 
We ablate Eq.~\ref{eqn:overall_energy} by removing affordance reasoning, grasp feasibility, or visual alignment. On the multi-grasp dataset, all removals reduce performance: KpNet drops from 55.67\% to 35.05\%, 41.24\%, and 47.42\%, respectively, and GraspNet drops from 20.62\% to 18.56\%, 18.56\%, and 16.49\%. Removing affordance reasoning causes the largest drop. Sum aggregation in \(E_{\mathrm{aff}}\) outperforms mean aggregation, 55.67\%/20.62\% vs. 50.52\%/17.53\% with KpNet/GraspNet, supporting summed evidence from multiple query-relevant properties. 
Varying the Gemini KG-generation temperature from 0.2--2.0 remains stable, with KpNet predictions ranging from 51.55--57.73\% and GraspNet from 17.53--22.68\%, indicating stability across LLM-derived priors.

\textbf{Qualitative Analysis.} 
Figure~\ref{fig:qualitative_results} illustrates the modular pipeline of CRAFT-E and highlights key failure modes. Grasp failures, often due to flat or thin objects like scissors or knives, result in the exclusion of functionally appropriate candidates from downstream reasoning. SAMv2, while enabling class-agnostic generalization, can oversegment tools into disjoint parts (e.g., blade and handle), leading to bounding boxes that fail the IoU threshold despite correct functional intent. Functional grounding errors often arise in cases of semantic ambiguity, where multiple objects plausibly afford the queried action, particularly under occlusion or uncertain pose. These examples underscore the value of CRAFT-E's modular design, which enables transparent attribution of errors across perception, embodiment, and reasoning stages. 
These failure modes primarily stem from upstream perception rather than the symbolic reasoning module itself. Grasp exclusion reflects the broader challenge of acting under embodiment constraints, where perceptual uncertainty suppresses valid candidates. Oversegmentation errors are a consequence of SAMv2’s class-agnostic nature, which lacks object permanence or part-whole consistency. Grounding failures reflect inherent ambiguities in object appearance and language, especially without contextual cues. 

\textbf{Navigating Semantic Drift.} 
A key distinction between LLM-derived affordance graphs and general-purpose knowledge graphs such as ConceptNet is whether paths are organized around functional relevance or generic semantic proximity. ConceptNet is broad and expressive, but its general-purpose relations can produce \emph{semantic drift}: paths may traverse linguistically related nodes that are not directly relevant to the queried function. For example, for \textit{write}, ConceptNet produces \texttt{write $\rightarrow$ RelatedTo $\rightarrow$ read $\rightarrow$ RelatedTo $\rightarrow$ writing $\leftarrow$ UsedFor $\leftarrow$ notebook}. Although each edge is plausible, the path explains \textit{notebook} through lexical association rather than its functional role as a writing surface. 
CRAFT-E instead constrains reasoning to an affordance-centric subspace by modeling intermediate functional properties. This yields property-centered paths such as \texttt{write $\rightarrow$ recordable $\leftarrow$ notebook}, which directly encode the object’s role in supporting the action. By grounding decisions in operationally relevant properties, CRAFT-E reduces ambiguity and avoids detours through unrelated concepts, leading to more interpretable rankings between related objects differing in functional use.

\begin{table}[t]
\centering
\resizebox{\columnwidth}{!}{%
\begin{tabular}{llccccccc}
\toprule
\textbf{Grasp Model} & \textbf{Model} & \textbf{3D} & \textbf{Detect} & \textbf{Grasp Rate} & \textbf{Case 1} & \textbf{Case 2} & \textbf{Case 3} & \textbf{Case 4} \\
\midrule
\multirow{4}{*}{GraspNet}
& Gemini  &  83.89 & 56.67 & 44.44 & 44.44 & 12.22 & 27.22 & 16.11\\
& GPT-4o   & 82.22 & 52.22 & 46.11 & 46.11 & 6.11 &  30.00 & 17.78\\
& CRAFT     & 86.11 & 37.78 & 33.33 & 33.33 & 4.44 &  48.33 & 13.89\\
& CRAFT-E  & 85.00 & 52.22 & 46.67 & 46.67 & 5.56 &  32.78 & 15.00\\
\midrule
\multirow{4}{*}{KpNet}
& Gemini & 72.78 & 45.00 & 23.89 & 23.89 & 21.11 & 27.78 & 27.22 \\
& GPT-4o & 78.89 & 45.00 & 33.33 & 33.33 & 11.67 & 33.89 & 21.11 \\
& CRAFT & 78.33 & 33.89 & 25.56 & 25.56 & 8.33 & 44.44 & 21.67 \\
& CRAFT-E & 77.22 & 51.67 & 40.00 & 40.00 & 11.67 & 25.56 & 22.78 \\
\bottomrule
\end{tabular}
}
\caption{\textbf{Real-world error analysis} across grasp backends and affordance priors (\%). Cases: (1) full success, (2) correct grounding/grasp but execution failure, (3) correct grounding but grasp failure, and (4) incorrect grounding.}
\label{tab:error_analysis}
\end{table}

\textbf{Error Analysis and Performance Bottlenecks.} 
To better understand the performance limitations of the full system, we analyze failures in real-world experiments by decomposing each trial into four categories: (1) full success, (2) correct grounding and grasp prediction but hardware execution failure, (3) correct grounding but incorrect or missing grasp prediction, and (4) incorrect grounding despite feasible grasp candidates. 
Results, summarized in Table~\ref{tab:error_analysis}, show that a significant portion of failures arises from perception and execution rather than affordance reasoning. For example, with CRAFT-E and GraspNet, 84 out of 180 trials result in full success, while an additional 10 cases involve correct grounding and grasp prediction but failure during physical execution. A larger fraction of failures (59 cases) correspond to incorrect grasp prediction despite correct object grounding, indicating that grasp generation is a dominant bottleneck. Similar trends are observed with KpNet. 
These findings suggest that the observed performance ceiling is primarily due to compounded uncertainty across perception, grounding, and physical execution, rather than limitations in the affordance reasoning module. This is further supported by static evaluation results under idealized perception, where accuracy increases to 64.95\%, highlighting the impact of upstream noise.
The GraspNet/KpNet comparison shows that backend-dependent failures can arguably be attributed to actionability. 

%% file: sec/6_conclusion.tex
\section{Conclusion and Future Work}

We introduced CRAFT-E, an interpretable and editable neuro-symbolic framework for affordance-grounded active object retrieval. By externalizing affordance reasoning into a verb--property--object graph, CRAFT-E connects functional language, visual evidence, and physical actionability to select objects that are both appropriate and retrievable. Across static, robotic, ImageNet, and COCO-Tasks evaluations, CRAFT-E provides a competitive and inspectable training-free prior for affordance-grounded retrieval, while enabling diagnosis and targeted graph edits without retraining. These results suggest that explicit affordance structure is useful not only for retrieval, but also for making embodied decisions inspectable and correctable. Current limitations stem from segmentation errors, unstable or unreachable grasps, limited contextual reasoning, and dependence on the coverage and quality of the generated affordance graph. Future work will explore instance-consistent perception, reachability-aware grasping, adaptive weighting, temporal interaction histories, automated edit discovery, held-out edit transfer, and targeted knowledge intervention for more robust and personalized retrieval. 

\textbf{Acknowledgement.} This research was supported in part by the US National Science Foundation grants IIS 2348689, IIS 2615771, CNS 2551658, and IIS 2348690, and the US ARO Award Number W911NF-26-2-A121.